\documentclass[a4paper]{article}

\usepackage[english]{babel}
\usepackage[utf8x]{inputenc}
\usepackage[T1]{fontenc}

\usepackage[a4paper,top=3cm,bottom=2cm,left=3cm,right=3cm,marginparwidth=1.75cm]{geometry}

\usepackage{amsmath}
\usepackage{graphicx}
\usepackage[colorinlistoftodos]{todonotes}
\usepackage[colorlinks=true, allcolors=blue]{hyperref}

\usepackage{longtable}
\usepackage{array}
\newcolumntype{L}[1]{>{\raggedright\let\newline\\\arraybackslash\hspace{0pt}}m{#1}}
\newcolumntype{C}[1]{>{\centering\let\newline\\\arraybackslash\hspace{0pt}}m{#1}}
\newcolumntype{R}[1]{>{\raggedleft\let\newline\\\arraybackslash\hspace{0pt}}m{#1}}
\usepackage{soul}

\title{Appendix - Recommended Statistical Significance Tests for NLP Tasks}
\author{Rotem Dror \hspace{1cm} Roi Reichart\\
Faculty of Industrial Engineering and Management, Technion, IIT\\
{\tt \{rtmdrr@campus$|$roiri\}.technion.ac.il}}
\date{}

\begin{document}
\maketitle

\noindent\fbox{%
    \parbox{\textwidth}{%
    This document is an annex to the paper "The Hitchhiker’s Guide to Testing Statistical Significance in Natural Language Processing".Rotem Dror, Gili Baumer, Segev Shlomov and Roi Reichart (ACL 2018). Please cite:\\
    \footnotesize{
@inproceedings\{dror2018hitchhiker,\\
\hspace*{3mm} title=\{The Hitchhiker’s Guide to Testing Statistical Significance in Natural Language Processing\},\\
\hspace*{3mm} author=\{Dror, Rotem and Baumer, Gili and Shlomov, Segev and Reichart, Roi\},\\
\hspace*{3mm} booktitle=\{Proceedings of the 56th Annual Meeting of the Association for Computational Linguistics (Volume \hspace*{3mm} 1: Long Papers)\},\\
\hspace*{3mm} volume=\{1\},\\
\hspace*{3mm} pages=\{1383--1392\},\\
\hspace*{3mm} year=\{2018\}\\
  \}}
    }%
}
\vspace{1cm}
\abstract{Statistical significance testing plays an important role when drawing conclusions from experimental results in NLP papers. Particularly, it is a valuable tool when one would like to establish the superiority of one algorithm over another. This appendix complements the guide for testing statistical significance in NLP presented in \cite{dror2018hitchhiker} by proposing valid statistical tests for the common tasks and evaluation measures in the field.}

\section*{Statistical Significance Test Table}
For each evaluation measure, the following table presents valid statistical tests\footnote{A statistical test is called valid if it guarantees that the probability of making type one error is bounded by a pre-defined constant. See \cite{dror2018hitchhiker} for a detailed explanation.} and explanations about the assumptions made when using each test.
Our recommendations are based on the considerations discussed in \cite{dror2018hitchhiker}. We will be happy to get feedback about their validity in case the reader does not found our considerations to hold.

Notice that for each measure we present both parametric and non-parametric tests that can be used under certain assumptions.
The parametric tests discussed here assume that the data is normally distributed. This assumption is likely to hold when using evaluation measures that calculate an average of counts of correct classifications. 
When the normality assumption that accompanies these tests holds in practice, they have higher statistical power than their non-parametric counterparts proposed in the table, hence it is recommended to use them. Otherwise, one should use non-parametric tests that do not make any such assumptions (we marked cases where it is unlikely to assume normality by --- in the parametric test column). 

For example, in the case of precision, recall and F-score, \cite{yeh2000more} described why the t-test can only be used for the recall metric but not for the precision and F-score measures. For other evaluation measures, one can test if the data is normally distributed by applying statistical tests that check for normality (see \cite{dror2018hitchhiker} for more details). In this table, we only mention a parametric test when we consider it likely to assume normality.

Additionally, when comparing the performance of two algorithms that are applied on the same dataset, one should use the paired version of the statistical significance test (such as the matched-pair t-test). An implementation of the paired versions of all statistical tests presented here as well as of other tests can be found at \url{https://github.com/rtmdrr/testSignificanceNLP}.

\newpage
\begin{longtable}{|L{2cm}|L{3.5cm}|L{2cm}|L{2cm}|L{2cm}|L{2cm}|}
\hline
\textbf{Evaluation Measure} & \textbf{Description} & \textbf{Parametric Test} & \textbf{Non-Parametric Test} & \textbf{Exemplary Task} & \textbf{Assumptions/ Comments} \\ \hline
Contingency table/ confusion matrix & $2\times 2$ table\footnote{\url{https://en.wikipedia.org/wiki/Confusion_matrix}} which presents the 
outcomes of an algorithm on a sample of $n$ data points: 
\begin{tabular}{l|l}
\# tp & \# fp \\ \hline
\# fn & \# tn \\
\end{tabular}
& --- & McNemar's test & Binary sentiment classification & \footnotesize{tp (true positive), fp (false positive), fn (false negative), tn (true negative)}  \\ \hline
Exact match & Percentage of predictions that match any one of the ground truth answers exactly & t-test & bootstrap/ permutation & Question answering & \ref{lab:ttest}, \ref{lab:bootstrap} \\ \hline
Accuracy & Proportion of true results (both true positives and true negatives) among the total number of cases examined\footnote{\url{https://en.wikipedia.org/wiki/Accuracy_and_precision}} & t-test & bootstrap/ permutation & Sequence labeling & \ref{lab:ttest}, \ref{lab:bootstrap} \\ \hline
Recall & $\frac{\mathrm{true\;positive}}{\mathrm{true\;positive} + \mathrm{false\;negative}} $\footnote{\url{https://en.wikipedia.org/wiki/Precision_and_recall}} & t-test & bootstrap/ permutation & Phrase-based (constituent) parsing & \ref{lab:ttest}, \ref{lab:bootstrap}, \ref{lab:more}\\ \hline
Precision & $\frac{\mathrm{true\;positive}}{\mathrm{true\;positive} + \mathrm{false\;positive}} $\footnote{\url{https://en.wikipedia.org/wiki/Precision_and_recall}} & --- & bootstrap/ permutation & Phrase-based (constituent) parsing & \ref{lab:bootstrap}, \ref{lab:more}\\ \hline
F score & $ F_\beta = (1 + \beta^2) \cdot \frac{\mathrm{precision} \cdot \mathrm{recall}}{(\beta^2 \cdot \mathrm{precision}) + \mathrm{recall}}$ & --- & bootstrap/ permutation & Semantic parsing & \ref{lab:bootstrap}, \ref{lab:more}\\ \hline
Perplexity & Measurement of how well a probability distribution or probability model predicts a sample\footnote{\url{https://en.wikipedia.org/wiki/Perplexity}} & --- & Wilcoxon signed-rank test& Language modeling & \ref{lab:Wilcoxon}\\ \hline
Spearman correlation & Measure of rank correlation between the ranking produced by two algorithms  \footnote{\url{https://en.wikipedia.org/wiki/Spearman\%27s_rank_correlation_coefficient}}& Z-test & bootstrap/ permutation & Word similarity & \ref{lab:bootstrap}, \ref{lab:fisher}\\ \hline
Pearson correlation & Measure of the linear correlation between the results of two algorithms\footnote{\url{https://en.wikipedia.org/wiki/Pearson_correlation_coefficient}} & Z-test & bootstrap/ permutation & Word similarity& \ref{lab:bootstrap}, \ref{lab:fisher}\\ \hline
UAS (sentence-level) & Unlabeled attachment score \cite{kubler2009dependency}& t-test & bootstrap/ permutation & Dependency parsing & \ref{lab:ttest}, \ref{lab:bootstrap}, \ref{lab:dependency} \\ \hline
LAS (sentence-level) & Labeled attachment score \cite{kubler2009dependency} & t-test & bootstrap/ permutation & Dependency parsing & \ref{lab:ttest}, \ref{lab:bootstrap}, \ref{lab:dependency} \\ \hline
ROUGE & \cite{lin2004rouge} & --- & bootstrap/ permutation & Summarization & \ref{lab:bootstrap} \\ \hline
BLEU & \cite{papineni2002bleu} & --- & bootstrap/ permutation & Machine translation & \ref{lab:bootstrap}\\ \hline
METEOR & \cite{banerjee2005meteor} & --- & bootstrap/ permutation & Machine translation & \ref{lab:bootstrap}\\ \hline
PINC & \cite{chen2011collecting} & --- & bootstrap/ permutation & Paraphrasing & \ref{lab:bootstrap} \\ \hline
CIDEr & \cite{vedantam2015cider} & --- & bootstrap/ permutation & Image description generation & \ref{lab:bootstrap} \\ \hline
% \# OOV & OOV=out of vocabulary & t-test & bootstrap/ permutation & & \ref{lab:ttest}, \ref{lab:bootstrap}\\ \hline
MUC, $B^3$, $CEAF_{e}$, BLANC & \cite{vilain1995model, bagga1998algorithms, luo2005coreference, lee2012joint} & --- & bootstrap/ permutation & Coreference resolution & \ref{lab:bootstrap}, \ref{lab:coreference}\\ \hline
Krippendorf's alpha, Cohen's kappa &  Statistical measures of agreement \cite{krippendorff2011computing, cohen1960coefficient} & --- & bootstrap/ permutation & Annotation reliability & \ref{lab:bootstrap}\\ \hline
MRR & Mean reciprocal
rank, used for algorithms that produce a list of possible responses to a sample of queries, ordered by probability of correctness\footnote{\url{https://en.wikipedia.org/wiki/Mean_reciprocal_rank}} & --- & bootstrap/ permutation & Question answering, information retrieval & \ref{lab:bootstrap}\\ \hline
\end{longtable}

\section*{Assumptions and Comments}
\begin{enumerate}
\item \label{lab:ttest}t-test: assuming that an average count of correct classifications is normally distributed.
\item \label{lab:bootstrap}Bootstrap: assuming the dataset is representative of the whole population. I.e., the data sample should be big enough to cover many use-cases and represent the domain in a satisfactory manner. 
\item \label{lab:fisher}Fisher transformation: $F(r) = \frac{1}{2}\ln\frac{1+r}{1-r}$.
\begin{itemize}
\item For Spearman: $\sqrt{\frac{n-3}{1.06}}F(r)$ approximately follows a standard normal distribution.
\item For Pearson: the Fisher transformation approximately follows a normal distribution with a mean of $F(\rho)$ and a standard deviation of $\frac{1}{\sqrt{n-3}}$, where $n$ is the sample size. One can use this to apply z-test to test for significance.
\end{itemize}
\item \label{lab:dependency} UAS and LAS are actually accuracy measures.
\item \label{lab:Wilcoxon} Wilcoxon signed-rank test: since the number of possible predictions (vocabulary words) a language model may make can be very large, the non-sampling non-parametric tests are preferable.
\item \label{lab:more} See \cite{yeh2000more} for explanations.
\item \label{lab:coreference} All measures used for evaluating coreference resolution are functions of precision and recall. Since parametric tests are not suitable in the case of precision, these tests are also not valid for these measures.

\end{enumerate}

\bibliographystyle{plain}
\bibliography{sample}

\end{document}